# **A linguistic evaluation of transcription errors produced by an automatic speech recognition system**

**Lauren Harrington[1], Jessica Wormald[1] & Tallulah Buckley[2]**

1 Department of Language and Linguistic Science, University of York, UK

2 Department of Theoretical and Applied Linguistics, University of Cambridge, UK

## *Abstract*

Automatic speech recognition (ASR) systems are increasingly employed to transcribe real-life spoken interactions. The outputs of these systems are known to contain errors, though developers often report only overall error rates which conceal vital information about the errors and their impacts on the transcript. The current article applies linguistic and phonetic analysis to errors produced by a state-of-the-art open-source ASR system with a dataset of thirteen male speakers with non-standard British English accents. We find that errors account for around 10% of the transcripts and that substitutions are the most common error type. Substitutions often involve content words such as verbs and proper nouns, and most are phonetically similar to the spoken content. The majority of substitutions do not retain the original meaning, and many do not make sense either grammatically or based on nonsense content. The article demonstrates the benefit of linguistic analyses in this domain, and concentrating on the erroneous content in ASR transcripts provides decision-makers with a better understanding of the performance of these systems and the risks that can be expected.

## *Introduction*

In this study, we employ linguistic and phonetic analysis to evaluate the performance of an automatic speech recognition (ASR) system with spontaneous speech. We explore (i) the types of errors produced by the system, and (ii) the impact of the errors with regard to meaning retention. The use of ASR is increasingly being adopted to transcribe interactions across a range of sectors, including high-stakes environments such as healthcare (Seuren, Williams & Cresswell 2026), probation (Ministry of Justice 2025) and law enforcement (Walley & Glasspoole-Bird 2025).

The evaluation of such technologies typically focuses on efficiency and time-saving measures. For example, the Ministry of Justice (2026) regularly publishes data about *Justice Transcribe*, a system that transcribes and summarises probation meetings. The data reports the total number of meetings summarised and an illustrative estimate of how much time has been saved, but no accuracy or quality-based metrics. However, there are also reports, such as those outlined in Sargeant (2026), which demonstrate that using AI-based speech technologies does not necessarily result in productivity gains and could inadvertently produce higher workload demands involving checking and correcting outputs.

Understanding ASR performance in realistic environments is crucial when deciding whether, when and where it is suitable for use in high-stakes contexts. However, the testing of ASR systems tends to be done on generic datasets and with numeric metrics which obscure nuance in system performance. The Open ASR Leaderboard (Srivastav et al. 2026), a popular online platform presenting live results from state-of-the-art ASR system testing, reports only two metrics: (i) word error rate (WER), which calculates the difference between the ASR transcript and a 'ground-truth' transcript, and (ii) inverse real-time factor, which calculates the speed with which the system processes the audio data. Such numeric metrics reveal no insight into the actual content of the transcripts, the types of errors the system produces and the impact of those errors on the content.

Without detailed and content-focused evaluation of ASR outputs, those looking to adopt this technology cannot assess system performance in a meaningful way. Risks and benefits therefore cannot be appropriately assessed, posing problems for high-stakes use of these technologies. This paper demonstrates how a linguistically-informed analytical approach can provide a better understanding of system performance by presenting explainable, accessible findings to inform decision-makers.

## *Background*

### *Automatic Speech Recognition*

Transcription is the process of representing spoken language in written form (Fraser 2022) and is a time-consuming procedure, often taking humans 4-10 times the duration of the audio to transcribe its content (Nagy & Sharma 2013). ASR systems automatically transcribe the speech within audio recordings or live spoken data, and modern state-of-the-art systems can often generate a transcript hundreds of times quicker than the duration of the speech (Srivastav et al. 2026). ASR systems traditionally consisted of three main components: an acoustic model (mapping audio to speech sounds), a pronunciation model (mapping speech sounds to written characters) and a language model (mapping character sequences to words and longer stretches of text) (Wang, Wang & Lv 2019). Modern end-to-end systems collapse those components into a single model that directly maps the acoustic signal onto word sequences, and are trained on huge datasets containing years' worth of audio data and reference transcriptions (e.g. Radford et al. 2023). In short, the ASR system uses information from the audio to determine the speech sounds that have been uttered, and uses its 'knowledge' of language (learned from the training data) to predict the most probable sequence of words that matches the identified speech sounds.

ASR models have shown rapid improvements in performance over the last two decades. In 2005, state-of-the-art ASR systems produced word error rates (WER) of around 20-30% (Hain et al. 2005), but the introduction of deep learning and neural networks has brought with it considerable improvements (Nayeem et al. 2025). The Open ASR Leaderboard (Srivastav et al. 2026) showcases the top-performing ASR systems, which currently achieve performance around 4.4% WER[1]. Systems are tested on a curated range of datasets including both read and spontaneous speech, for which the top performing ASR system achieves error rates of around 2% and 5-12% respectively. Spontaneous speech is a known challenge for ASR systems (Jurafsky & Martin 2026), and many other factors have been found to negatively impact performance. Poor performance is often attributed to underrepresentation in the training data (e.g. Markl 2022). These factors include:

- Speech-related factors such as slang terminology (Klering et al. 2026) and emotional speech (Tuttösí et al. 2026)
- Speaker-related factors such as age (Feng et al. 2024), ethnicity (Koenecke et al. 2020) and accent (Markl, 2022)
- Environment-related factors such as noise (Dua, Akanksha & Dua 2023), overlapping speech (He & Whitehall 2026) and distance from microphone (Tuttösí et al. 2026)

*Evaluation of ASR outputs*

To evaluate ASR performance, the transcript produced by the system is compared with a ‘ground truth’ or ‘reference’ transcript that has been created by a human. Word error rate (WER) is the standard metric used to evaluate ASR system performance (Jurafsky & Martin 2026) and is calculated by dividing the total number of errors (i.e. differences between the transcripts) by the total number of words in the reference transcript. Transcripts are aligned on a word-level, and word pairings are marked as correct or as an error. Errors fall into three categories: substitutions (where the words in both transcripts do not match), deletions (where there is no word in the ASR transcript), and insertions (where there is no word in the reference transcript). WER treats all three error types as equal, such that transcripts 1-3 in Table 1 would all score the same WER (25%). Particularly in high-stakes contexts like policing and healthcare, the mistranscription of a single word could have a serious impact. For example, the change in pronoun (Transcript 1) and negation (Transcript 3) could impact the direction of a police investigation, or later be used as evidence against the interviewee (see Haworth 2018).

| **Reference transcript** | **Transcript 1 (SUB)** | **Transcript 2 (DEL)** | **Transcript 3 (INS)** |
|---|---|---|---|
| I was at home | **He** was at home | I was home | I was **not** at home |

*Table 1: Examples of different transcriptions of the utterance “I was at home”, each containing one error.*

Transcription is a subjective process which is impacted heavily by decisions that the transcriber makes, both consciously and unconsciously (Bucholtz 2000). A difference between transcripts may demonstrate different ways to represent the speech (e.g. “gonna” versus “going to”), but a word-level comparison of transcripts will judge all differences in the same way. To address the disadvantages of WER, ASR evaluation may involve alternative metrics like SemDist (Kim et al. 2021) and BERTScore (Zhang et al. 2019), which convert text into high-dimensional vector embeddings and calculate the cosine similarity between those embeddings. These metrics attempt to measure the similarity in meaning between two units (sentences or words respectively). However, the numerical outputs can be difficult to interpret, and the impact of transcription errors largely depends on the context within which the error is found. While two sentences may be similar from the perspective of computational semantics, the human understanding and perception of those sentences may differ.

Employing linguistically-informed approaches to evaluating ASR performance can help to explain why transcription errors come about, i.e. what is it about the speech content that is challenging for the ASR system. This enables decision-makers to have a better understanding of ASR performance and, in turn, be better informed of the risks. A small but growing body of work has taken a more error-focused approach to evaluating transcription performance, e.g. Harrington (2023). Alongside WER, the study presented the total number of errors and the number of each error type (substitution, deletion, insertion) produced by the ASR systems tested. Additionally, word groups (function and content words) were explored to better understand the type of content involved in mistranscriptions. WER results revealed an imbalance in performance across two varieties of British English, and the error analysis expanded on this, showing both *how* and *why* performance was worse for the speakers of the non-standard regional variety (West Yorkshire English; WYE). The paper demonstrated that (i) the majority of errors for the WYE speakers were substitutions (compared with mostly deletions for the speakers of Standard Southern British English; SSBE), (ii) these substitutions were often phonologically motivated, and (iii) the majority of substitutions for the WYE speakers were related to content words, many of which resulted in nonsense (e.g. "I had a bit of **dessert**" transcribed as "I had a bit of **Giza**").

To further explore the impact of substitution errors, Harrington (2024) attempted to create a substitution error categorisation framework, assigning substitutions into 'minor' or 'potentially major' categories. Substitutions involving synonyms and small morphological or grammatical alterations were categorised as 'minor' in their impact on the transcript's content, whereas all other substitutions were marked as having a 'potentially major' impact. Two issues with this approach are that (i) the analysis was done on word-level comparisons, disregarding errors which span multiple words, and (ii) the majority of errors fell into the 'potentially major' category. Further categorisation and manual analysis would be required to determine the actual impact of each of those errors on the transcript, but was out-of-scope for the study.

Serditova, Tang & Steffens (2025) employ an alternative linguistically-informed manual approach to error analysis. They focus mostly on substitutions and define ASR errors as either phonological, lexical, morphosyntactic, or resulting from standardisation or misspelling. Using Newcastle English, a regional variety of British English that diverges significantly from SSBE, the paper demonstrates that the majority of errors are related to phonological features, namely glottalisation, monophthongisation and vowel quality. Similar phonology-related findings for West Yorkshire English were demonstrated in Harrington (2023). The paper also highlights the value of sociolinguistic analysis in diagnosing issues with ASR performance related to accents, a sentiment shared by Wassink et al. (2022), who suggest that such findings can improve training data collection processes and promote inclusivity of speech technologies.

*Use of ASR outputs*

The outputs of ASR systems can be used in several ways. The transcript itself may be the final product or it may be used as the foundation for further processing such as summarisation and creation of structured reports or statements. It is widely acknowledged within high-stakes public sector contexts that AI outputs must be checked by a human (e.g. College of Policing 2026), often referred to as 'human-in-the-loop' or 'human verification'. This safeguard is extremely important in scenarios where one error could make a significant difference to, for example, a police investigation or a healthcare decision.

However, a major concern from a linguistic perspective is 'priming', whereby exposure to a stimulus can subconsciously influence listeners to hear certain words (Warren 1970). Previous work focusing on forensic audio has demonstrated that both contextual information about the recording situation (Lange et al. 2011) and exposure to an existing transcript (Fraser & Kinoshita 2021) can have a significant impact on listeners' perceptions of the speech content. This is particularly concerning given that the main safeguard considered for ASR in high-stakes contexts is human post-review of transcripts. Yet research has demonstrated the variable efficacy of human verification procedures, including overreliance on machine outputs and comprehensiveness of review procedures (e.g. Green 2022, Preston et al. 2026). Further, many use cases for ASR involve automatic generation of a summary, for which the verification process may involve only the summary (e.g. Justice Transcribe; National Audit Office 2026).

However ASR is being used, whether to generate verbatim transcripts of spoken interactions or as the basis for summarisation, being equipped with a linguistically-informed understanding of the performance of ASR systems is incredibly beneficial. Knowing how systems work and what types of factors are likely to impact performance, as well as being aware of the typical errors found in ASR transcripts, will enable those working in high-stakes contexts to make more informed decisions around the specific speakers, topics and use cases this technology could and should be used for. Furthermore, the design and development of appropriate human verification procedures cannot be facilitated without knowledge of these areas.

## *Aims*

The general objective of this study is to provide linguistically-informed insights about the performance of a state-of-the-art ASR system. This study expands on the work of Harrington (2023, 2024), employing their novel transcription error evaluative framework, whereby different types of errors and their implications are explored. In comparison with Harrington (2023), the present study utilises a larger sample of speakers from a similar demographic (i.e. young male speakers of British English), explores an additional regional accent (East Midlands English in addition to West Yorkshire English), and investigates ASR performance across longer stretches of speech. Furthermore, we enhance the evaluation of substitutions conducted in Harrington (2024) by investigating additional factors such as meaning retention, and by considering the context surrounding the error (rather than just the error itself). To address these areas, we pose two research questions:

1. What types of transcription errors are produced by the ASR system?
2. To what extent do substitution errors retain the original meaning of the reference transcript?

## *Methods*

### *Recordings*

We used recordings from the Punjabi English in Bradford and Leicester corpus (Wormald 2016), which includes English-speaking British-born individuals from Bradford and Leicester. We used a subset of thirteen young male speakers aged between 19 and 29. These speakers represent a single broadly-defined demographic (young British male) divided into two non-standard accent groups (West Yorkshire English and East Midlands English) and two language backgrounds (Anglo and Punjabi). Table 2 includes pseudonyms for the individuals included in this study, grouped by language background and city.

| **Language background / City** | **Bradford** | **Leicester** |
|---|---|---|
| **Anglo English** | Jake, Matthew | Andy, Simon |
| **Punjabi English** | Halim, Hamza, Irfan | Adeep, Hanspal, Hardev, Jasbir, Nihal, Satbir |

*Table 2: Pseudonyms of participants from each language background and city.*

Speakers took part in a paired-conversation task, discussing topics such as daily life, hobbies and cultural heritage with another participant, and were recorded in a quiet room using individual Beyerdynamic TG H54c neck-worn microphones. Single-speaker extracts were manually prepared and limited to five minutes in duration (301-310 seconds) to facilitate comprehensive qualitative analysis. We used samples containing the speech of individuals as ASR is known to struggle with speaker diarisation and attribution in multi-speaker recordings (e.g. Reid et al. 2026). In sum, the stimuli represent good technical quality audio containing spontaneous speech from a single speaker, creating an environment where we would expect relatively good ASR performance.

*Transcription*

The ASR transcripts were generated using CrisperWhisper[2], a variant of Open AI's Whisper (Radford et al. 2023). Whisper is a popular open-source ASR model that is used as the underlying model in many popular applications, e.g. ChatGPT (OpenAI 2023). We chose CrisperWhisper given that it claims to substantially mitigate hallucinations, significantly improve word-level time stamps and achieve "near perfect filled pause detection accuracy" (Wagner et al. 2024: 1). The JSON output provided by CrisperWhisper was converted to a Praat TextGrid file (Boersma & Weenink 2025), providing a word-level time-aligned transcript.

To produce the reference transcript, the output from CrisperWhisper was carefully reviewed and corrected by each author sequentially. This was done by the third author and second author in turn, followed by several rounds of review by the first author before the reference transcript was finalised. Given our interest in the lexical content of the ASR transcripts, we did not adjust word-level timestamps or amend punctuation, though these are both areas in which several errors were observed (example shown in Figure 1).

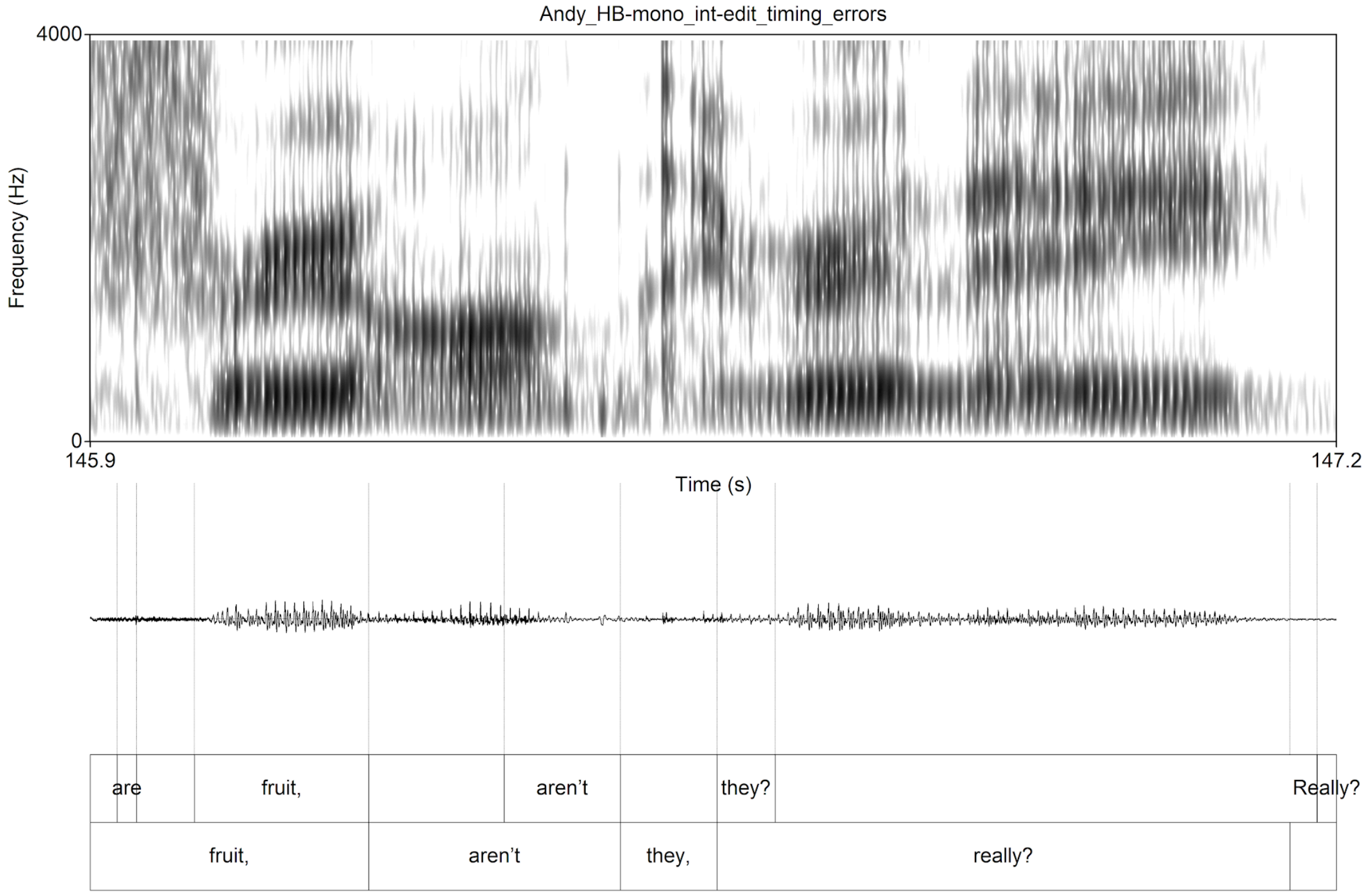

*Figure 1: Example of errors in ASR time alignment as demonstrated in Praat, whereby the first row of text utilises the timestamps provided by CrisperWhisper and the second row contains a manually-aligned transcription of the speech content. Note that the ASR transcript also contains inaccurate punctuation (question mark after "they" despite its utterance-medial position).*

There were a number of occasions where we considered the speech to be unintelligible, i.e. it was not possible to arrive at a reliable interpretation of the spoken content. The ASR system offered a written interpretation for 57/69 unintelligible sections (83%). For the remaining 17%, it did not attempt to transcribe the speech but included no indication of having excluded content. We have not included these utterances in our analyses, but this is an area in which further work would be particularly valuable.

*Quantifying errors*

Using the aligned reference and ASR transcripts from the Praat TextGrids, word pairings were categorised as either correct or as an error (i.e. a difference between the transcripts). Transcription errors were categorised as either a substitution (different content in the ASR and reference transcripts), a deletion (content missing from the ASR transcript which is present in the reference transcript), or an insertion (additional content in the ASR transcript not present in the reference transcript). We adapted the typical approach to error quantification and allowed single substitution errors to span multiple words in order to avoid artificially inflating the number of errors. For example, "**summer's**" was transcribed by the ASR system as "**some of us**". A traditional WER calculation would interpret this mistranscription as 3 separate errors (1 substitution and 2 insertions). Instead, we analyse this as a single substitution error, given that the ASR transcript contains an alternative interpretation of the same phonetic content. The stimuli and error quantification employed in this study also feature in Wormald & Harrington (2026), which explores the impact of features associated with spontaneous speech on by-speaker ASR performance.

*ASR transcription errors*

We calculated the number of each error type (substitution, deletion, insertion) per speaker and these were assigned to lexical categories according to the reference word (or the ASR-transcribed word in the case of insertions, where there is no reference word). This provides an overview of the word classes that are commonly mistranscribed, omitted or inserted into ASR transcripts. The lexical categories included were: adjective, adverb, auxiliary verb, conjunction, determiner, discourse marker (e.g. "yeah"), disfluency (e.g. part-word units), filled pause, modal verb, noun, number, preposition, pronoun, proper noun, and verb.

*Substitution analysis*

We are particularly interested in substitution errors, i.e. words or phrases that have been incorrectly transcribed by the ASR system (see Harrington 2023). We manually evaluated each substitution error according to three factors related to impact on the meaning within the content of the transcript and potential difficulties with checking the accuracy of the transcript (i.e. if the errors make sense and sound similar). We appreciate that there is a level of subjectivity to the approach, but consider this to be a necessary first step in developing appropriate frameworks for assessing the impact of errors with high-stakes applications in mind.

First, we evaluated whether the original meaning was retained, taking into account the surrounding content and context. 'Errors' are differences between the human-produced reference transcript and the ASR transcript. However, the ASR system representing something differently (e.g. "gonna" transcribed as "going to") or standardising grammar (e.g. "they was" transcribed as "they were") will not have the same level of impact as other substitutions (e.g. "gum" transcribed as "gun"). We therefore quantify whether the meaning has been retained by employing a binary classification (examples shown in Table 3).

| | **Reference transcript** | **ASR transcript** |
|---|---|---|
| **Meaning retained** | "Something **along** those lines" | "Something **on** those lines" |
| **Meaning not retained** | "Do you listen to any sort of music **or radio**?" | "Do you listen to any sort of music **already**?" |

*Table 3: Examples of substitution errors where meaning is retained (row 1) or **not** retained (row 2). In each case, the erroneous part is highlighted in bold, underlined text.*

Second, we evaluated whether the substitution made sense within the surrounding context, which was a holistic judgement based on nonsensicality and grammaticality. Harrington (2023) demonstrated that ASR errors can result in nonsense transcriptions, which broadly fall into two categories: content-related (e.g. "giving him an **haircut**" transcribed as "giving him an **airport**") and grammar-related (e.g. "I don't **know**" transcribed as "I don't **you**"). ASR models work by essentially predicting the most likely sequence of words, which means that incorrect transcriptions may often contain perfectly grammatical constructions. We quantify nonsensicality using a binary classification (examples shown in Table 4).

| | **Reference transcript** | **ASR transcript** |
|---|---|---|
| **Makes sense in context** | "**What I** would change **about** hometown" | "**Why** would **we** change **the** hometown" |
| **Does not make sense in context** | "You **support Leicester**" | "You **sport less time**" |

*Table 4: Examples of substitution errors that either do (row 1) or do **not** (row 2) make sense while reading the transcript. In each case, the erroneous part is highlighted in bold, underlined text.*

Third, we evaluated the degree of phonetic similarity between the words in the reference transcript and ASR transcript. Given that ASR models calculate the most probable sequence of words matching the speech sounds identified, many substitution errors produced by the ASR system will likely be phonetically similar to the spoken content. We quantified the **degree of phonetic similarity** on a three point scale where 0 means little to no similarity, 1 demonstrates some phonetic similarity (i.e. shared phonemes and comparable syllable structure), and 2 represents identical phonetic content (e.g. homophones). Examples are shown in Table 5.

| | **Reference transcript** | **ASR transcript** |
|---|---|---|
| **0: Little to no phonetic similarity** | "For **disabled** kids" | "For **several** kids" |
| **1: Some phonetic similarity** | "It's just a **prompt**" | "It's just a **problem**" |
| **2: Identical phonetic content** | "They're **Sikh**" | "They're **seek**" |

*Table 5: Examples of substitution errors according to phonetic similarity with the word in the reference transcript, with increasing phonetic similarity from row 1 to row 3. In each case, the erroneous part is highlighted in bold, underlined text.*

## ***Findings***

### *Error type analysis*

Figure 2 shows the total number of errors in the ASR transcripts for each error type. In total, there were 1341 errors across the dataset, which represents around 10% of the content of the ASR transcripts. Substitutions accounted for the majority of errors within the ASR transcripts, representing 59% of the total errors. Content words (adjectives, adverbs, nouns, numbers, proper nouns and verbs) accounted for almost half (48.7%) of all substitution errors. Function words accounted for 41.4%, and disfluencies or discourse markers accounted for 9.9%. The most common word classes that were substituted were pronouns (17.3%), nouns (13.7%) and verbs (13.1%).

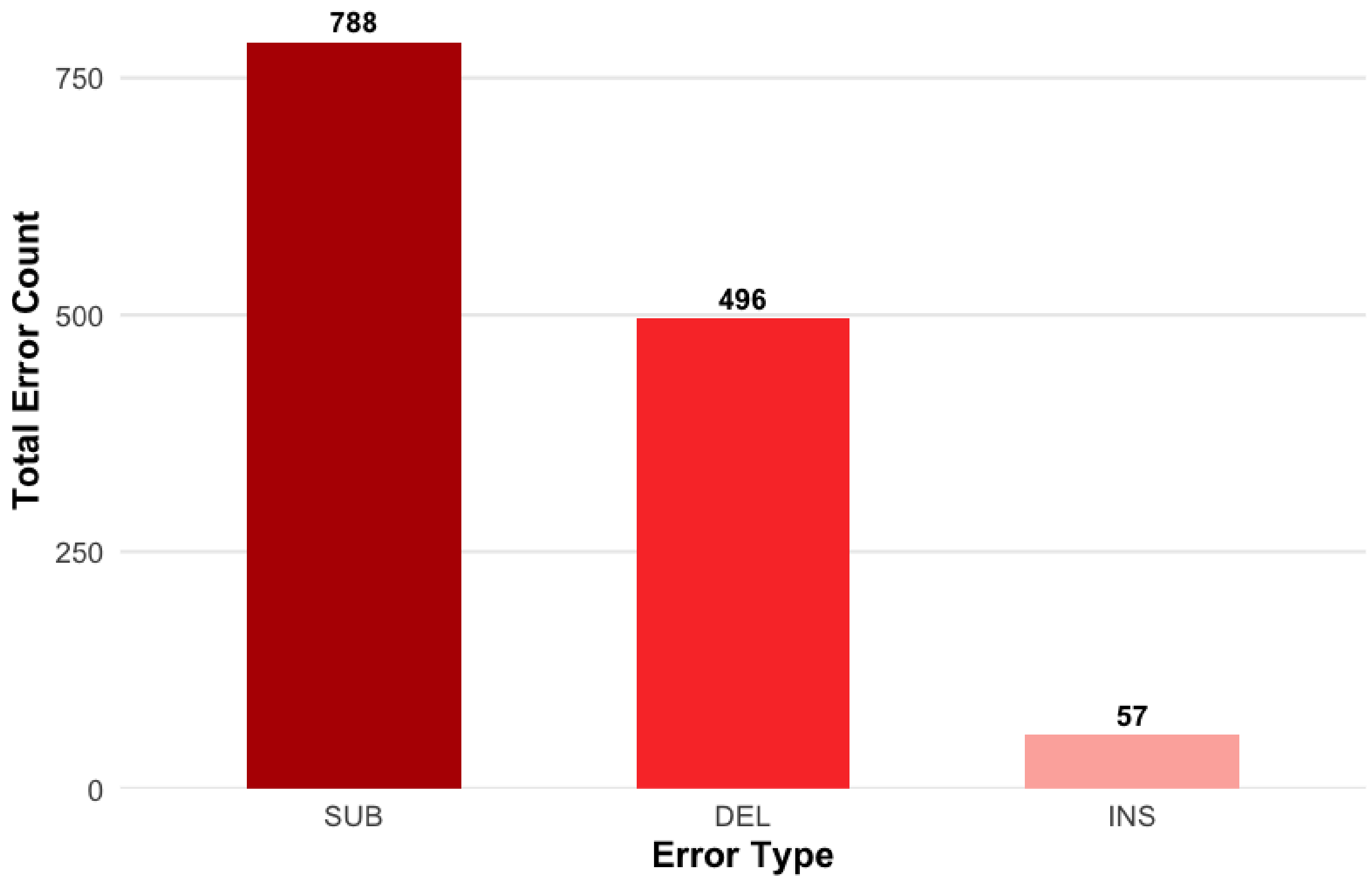

*Figure 2: Total error counts per error type, with counts included above each bar. Error types include substitutions (SUB), deletions (DEL), and insertions (INS).*

Interestingly, we observed that a notable proportion (7.4%) of substitutions involved proper nouns, e.g. place names and religious terminology (see examples in Table 6). The ASR system often transcribed these with variable accuracy, e.g. "Bradford" and "Leicester" were transcribed correctly 12/16 times and 8/13 times respectively. A similar pattern of inconsistency was observed with Sikh terminology, such as "Gurdwara" (accurately transcribed 1/3 times), indicating that the system had the capacity to correctly transcribe these terms, yet did not achieve complete accuracy.

| | **Reference transcript** | **ASR transcript** |
|---|---|---|
| **British place names** | "Barnoldswick" | "battles were considered" |
| | "Bradford" | "Baffin", "Brazil", "profile" |
| | "Eccleshill" | "a cool little" |
| | "Leicester" | "less time", "less to", "let's" |
| **Religious terminology** | "Akalis" | "colleagues" |
| | "Gurmukhi" | "good monkey" |
| | "Khalsa Panth" | "Khartoum" |
| | "Siri Hazur Sahib" | "silly has also" |

*Table 6: Examples of substitution errors involving proper nouns related to British place names and religious terminology.*

Deletions were the second-most frequently produced error type, representing around 37% of errors in the ASR transcripts. Function words were involved in the majority (55.6%) of deletion errors, while content words and those related to disfluencies or discourse markers accounted for 23.8% and 20.6% respectively. The most common word class to be omitted from transcripts was pronouns (27.2%), and determiners, disfluencies and adverbs each accounted for around 10% of deletions (11.3%, 10.9% and 9.1% respectively). The 'disfluencies' word class included only filled pauses and word-level repetitions and interruptions, but when combined with those at the phrase-level, it was found that around 30% of deletions occur at points of disfluency. Regarding deletions that involved multiple words, there was one instance where eleven seconds of speech (around 50 words) was not transcribed by the ASR system, and several other deletions involved consecutive units (e.g. "haven't had the **thrill of putting lights on**" transcribed as "haven't had the").

Insertions were rare within the data (around 4% of all errors), aligning with other studies on ASR error types (e.g. Harrington 2023). The vast majority of insertions (79.0%) were function words (pronouns, determiners, prepositions, auxiliary verbs, discourse markers and conjunctions). Many of these clearly resulted from the ASR system 'grammaticalising' the speech content. Speech sounds in function words often get elided (Jurafsky et al. 2001), resulting in utterances that are not wholly grammatical. Many of the insertions observed in our data functioned in this way (e.g. "for long time" transcribed as "for **a** long time"). There were several other instances where insertions were likely the result of a phrasal interruption (examples shown in Table 7). In these cases, the ASR system favours a grammatical sentence over an accurate reflection of the speech sounds, resulting in varying levels of meaning retention. Hallucinations involve longer stretches of inserted materials that are phonetically and semantically unrelated to the spoken content (Atwany et al. 2025) and are a known issue for Whisper models (e.g. Barański et al. 2025). We found only one hallucination-like error in our data, in which a fifteen-word utterance was erroneously transcribed twice.

| **Reference transcript** | **ASR transcript** |
|---|---|
| "I don't know- pop culture is-" | "I don't know **what** pop culture is" |
| "You can't see- send uh read receipts" | "You can't see **him** send uh read receipts" |

*Table 7: Examples of insertion errors that occur during phrasal interruptions. In each case, the erroneous part is highlighted in bold, underlined text.*

*Substitution analysis*

Figure 3 shows all substitution errors in the dataset categorised according to (i) whether the original meaning was retained, (ii) whether the error makes sense within the surrounding context, and (iii) how phonetically similar the reference and ASR words are. Initial observations demonstrate that the vast majority of substitution errors did *not* retain the original meaning of the utterance (88.5%) and that substituted words made sense around half (50.6%) of the time.

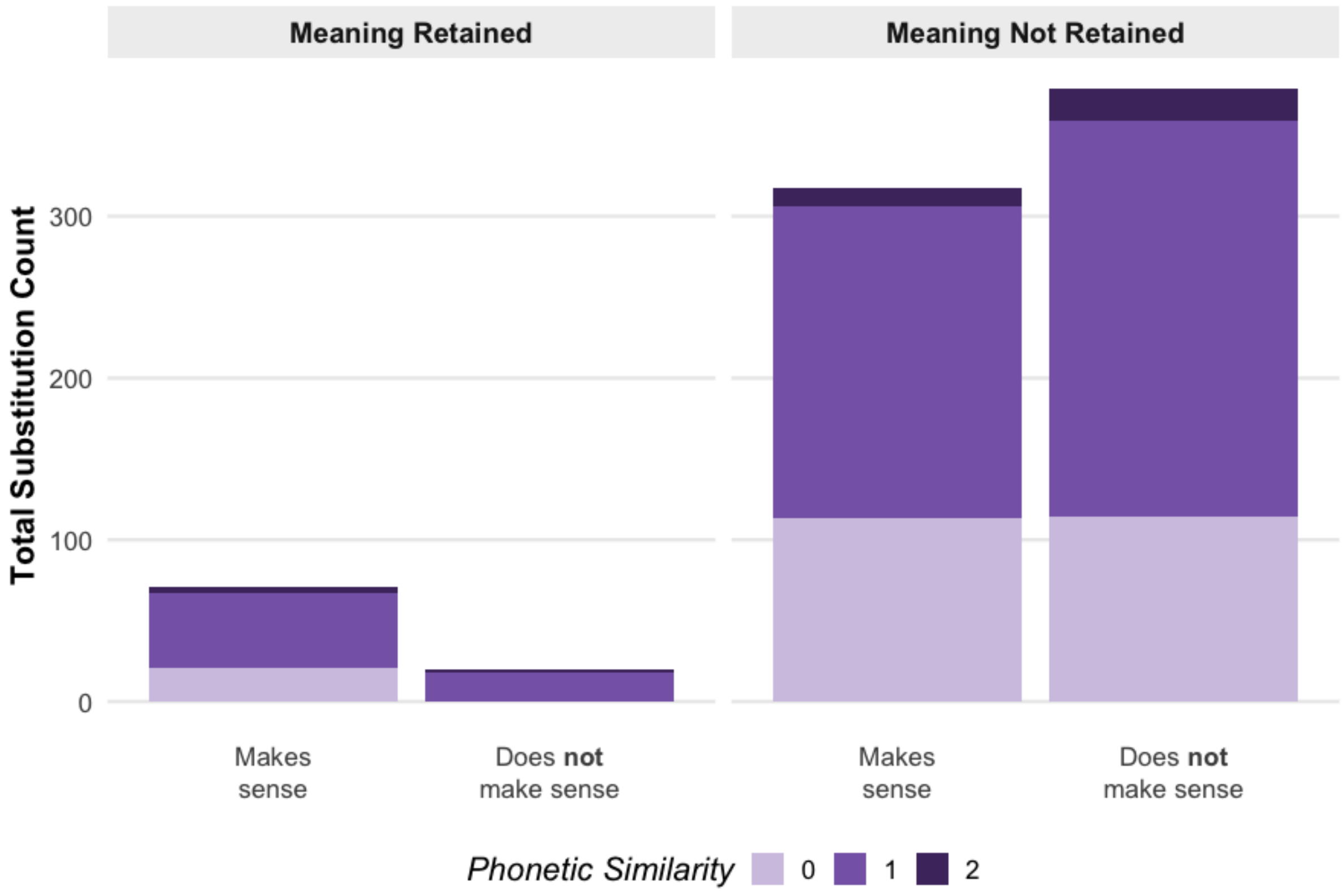

*Figure 3: Total number of substitutions within the dataset, categorised according to meaning retention and nonsensicality, and colour-coded according to the level of phonetic similarity (darker colours represent a higher level of phonetic similarity).*

Of particular interest is how these factors interact, e.g. how many of the substitutions which did *not* retain the original meaning still made sense within the surrounding context. Of the small number of substitutions where the original meaning was retained, most of these made sense within the transcript (e.g. the tag question in “all gotta learn at some point, **haven’t** we?” transcribed as “**don’t** we?”). There was also a small number of these substitutions which did *not* make (grammatical) sense, an example of which is the utterance “it’s **Pakistan** versus England" which was transcribed as “it’s **Pakistani** vs England” (i.e. replacing a noun with an adjective when the predictable structure is “country X versus country Y”).

Conversely, of the much larger number of substitutions where the original meaning was *not* retained, the majority of these did *not* make sense within the surrounding context (54.4%) due to unexpected grammatical constructions or vocabulary. Table 8 shows examples of substitution errors where meaning had *not* been retained; the top half of the table presents substitutions which did *not* make sense within the context (with increasing levels of nonsensicality), and the bottom half presents substitutions which did make sense.

| | **Reference transcript** | **ASR transcript** |
|---|---|---|
| **Does not make sense** | “For someone like him to do **something** like this is unbelievable” | “For someone like him to do **someone** like this is unbelievable” |
| | “It says that **you’re not, I’ve** not either” | “It says that **even if** not either” |
| | “They’ve proven to be cheaper than the top six, which includes British **<u>Gas, NPower</u>**, Eon” | “They’ve proven to be cheaper than the top six, which includes **<u>British cars and power</u>** Eon” |
| | “There is a different **<u>Bra- Bradford</u>** accent than to a Birmingham **<u>accent-</u>** accent” | “There is a different **<u>ba Baffin</u>** accent than to a Birmingham **<u>absent</u>** accent” |
| **Makes sense** | “It’s just **<u>where</u>** you live” | “It’s just **<u>the way</u>** you live” |
| | “When**<u>’s he coming</u>**?” | “When **<u>did you come in</u>**?” |
| | “I **<u>did</u>** look at that” | “I **<u>didn’t</u>** look at that” |
| | “You can *get Ur-* do **<u>Urdu</u>** [lessons] in **<u>Bradford</u>**” | “You can do **<u>it</u>** in **<u>Brazil</u>**” |

*Table 8: Examples of substitution errors that do **not** retain the original meaning. Examples in the top 4 rows do **not** make sense in context, and the bottom 4 rows include examples which do make sense. In each case, the erroneous part is highlighted in bold, underlined text.*

Overall, the majority of substitutions were phonetically similar to the reference transcript word. 63.7% shared some phonetic similarity and 4.8% were phonetically identical (i.e. homophones). Less than a third of the substitution errors (31.5%) were *not* phonetically similar to the word in the reference transcript. These findings align with our expectations given that ASR systems identify the speech sounds within the recording and, based on that information, predict the most probable sequence of words. In cases where the ASR transcription deviated from the speech sounds, a likely explanation is that the ASR system favours a grammatical and predictable sentence over fidelity to the phonetic content (as discussed in McGuire & Larson-Hall 2025). Many of the ‘phonetically dissimilar’ substitutions involved (i) non-standard grammatical constructions, (ii) disfluencies, and (iii) nearby phonetically-similar errors (examples shown in Table 9).

| | **Reference transcript** | **ASR transcript** |
|---|---|---|
| **Grammar** | “I **<u>were</u>** like- just messing about” | “I **<u>don’t</u>** like just messing about” |
| **Disfluency** | “Go to **<u>their f-</u>** house” | “Go to **<u>have a</u>** house” |
| **Nearby error(s)** | “What *other* interests **<u>do</u>** you have?” | “What *are the* interests **<u>that</u>** you have?” |

*Table 9: Examples of substitution errors that share no phonetic similarity with the word in the reference transcript. In each case, the erroneous part is highlighted in bold, underlined text.*

Most substitutions which (i) do not retain the original meaning but (ii) make sense in context share a high degree of phonetic similarity with the corresponding reference transcription. Over a quarter of mistranscribed words (26.5%) ‘look right’ (i.e. they make sense, are grammatical and do not immediately register to a reader as wrong) and ‘sound right’ (i.e. containing many or all of the sounds spoken) but do not retain the original meaning. This reveals that a substantial number of substitutions within ASR transcripts will likely be challenging to identify within a human verification process.

## ***Discussion***

In this study, we explored the performance of an open-source ASR system on single-speaker samples of spontaneous speech from young male speakers of British English. Although the majority of speech content was transcribed correctly for most speakers, all transcripts contained at least some substantial errors. We have employed linguistic analysis to improve our understanding of when and why errors occur in ASR transcripts. Our findings can inform both the development and refinement of ASR systems, and how organisations identify appropriate contexts and necessary safeguards for safe use of ASR.

*Research Question 1: What types of transcription errors are produced by the ASR system?*

All of the transcripts produced by the ASR system contained errors, and these accounted for around 10% of the transcript content for this dataset. Overall, substitutions were the most common type of error that the ASR system produced, accounting for around 60% of all transcription errors. Words from all word classes were incorrectly transcribed within our dataset, with around 49% of substitutions involving content words, 41% involving function words, and 10% involving disfluencies or discourse markers. These findings mirror those in Harrington (2023), where the majority of ASR errors for speakers of West Yorkshire English were substitutions, and that substitutions for those speakers were most commonly related to content words.

Pronouns were the most commonly substituted word class, and these substitutions often featured within sections of consecutive errors (e.g. "**he's** getting towards **his** thirties" transcribed as "**it's** going to all just **us** that"). It is typical for pronouns to be phonetically reduced in spontaneous speech (e.g. Jurafsky et al. 2001), often produced as a single unstressed vowel. This limits the acoustic information from which the ASR system can predict the content, hence it is unsurprising that errors frequently occur at these points. There were also seventeen examples of pronouns being swapped within otherwise correctly transcribed sections, resulting in a change of person (e.g. "both of **us**" transcribed as "both of **you**"). Almost all of these pronoun-pronoun substitutions resulted in the meaning not being retained.

Nouns and verbs accounted for just over a quarter of substituted words and only 7.5% of these substitutions retained the original meaning. Proper nouns, which carry important lexical meaning, were transcribed with variable accuracy by the ASR system. For example, "Bradford" and "Leicester" were correctly transcribed on some occasions but not others. Additionally, some other common British terminology was incorrectly transcribed by the system (e.g. "GCSE" transcribed as "GCC" and "to see"). Thus, whether proper nouns are recognisable for a given ASR system will not be consistent or necessarily predictable. Names, locations, medical conditions and drug names will feature frequently across a range of high-stakes contexts such as healthcare and policing. Those considering adopting the technology must recognise the risks of mistranscriptions of proper nouns and build safeguards to mitigate the impact.

We did not systematically evaluate deletion and insertion errors for their impact on the meaning of the transcript's content or whether they made sense within the given context. Deletions accounted for just under 40% of all errors and insertions were rare (4%). Both error types most commonly involved function words and disfluencies, with around 30% of deletions occurring at points of disfluency. Insertions either standardised grammar, occurred alongside other error types, or added disfluencies. Many deletions and insertions seemed to be inconsequential, but there were examples of both which resulted in a significant change in meaning (e.g. "York" transcribed as "**New**

York”). We focused our analysis on substitutions given their propensity for meaning change and phonetic similarity, but all error types have the capacity to significantly impact the content of the transcript and future work should aim to explore all types of transcription errors.

By classifying the errors into the different types (substitutions, deletions and insertions), we can better understand the content of the transcripts. Verbs and nouns account for a substantial proportion of substitutions, and proper nouns can cause problems. We found that almost a third of deletions involved disfluencies, though the impact of this type of error depends on the purpose of the transcript. If the transcript is designed to be an easily readable written record of the speech, then the deletion of disfluencies may not be viewed as an issue. However, if the transcript’s purpose is to present a strictly verbatim record of the speech content, the deletion of any content is unfavourable. The purpose of the transcript and typical amendments that the model makes to spoken language (e.g. removing disfluencies or standardising grammar) should be taken into consideration when considering the use of automatic transcription.

*Research Question 2: To what extent do substitution errors retain the original meaning of the reference transcript?*

In this study, we expanded on previous work by the first author on the categorisation of substitution errors according to their impact on the transcript. The analysis in Harrington (2024) was conducted systematically on a large-scale dataset with broad labels for ‘minor’ and ‘potentially major’ substitution errors. By employing manual evaluation of substitutions in the present study, we have been able to explore mistranscriptions within the ASR transcripts in a more nuanced and detailed manner.

We found that the majority of substitution errors do not retain the original meaning, although this does not necessarily correspond with a *significant* change in meaning. For example, the statement “I **didn’t** know that” was transcribed as “I **don’t** know that” by the ASR system. This results in a change in tense, whereby information that was not known at a previous point in time transforms into information that is not known at the time of the utterance. The significance of the substitution will depend entirely on other contextual information.

We found that around half of all substitutions resulted in nonsense transcriptions, which could broadly be separated into those related to (i) content and (ii) grammar. Additionally, the majority of substitutions where the original meaning was not retained resulted in nonsense transcriptions. The potential impact of these nonsense transcriptions on automatically generated summaries is also an area of particular importance for future research. For many applications currently being developed, ASR of spoken content is the first step in a more complex process. Individuals, organisations and developers should ensure appropriate safeguards are in place to mitigate the risk of nonsense utterances becoming embedded into summaries and reports.

The majority of substitution errors in the data were phonetically similar to the reference content, i.e. the reference and ASR transcriptions shared a number of speech sounds or sounds with similar phonological features. There were 38 occasions where the ASR system transcribed a homophone or a homophonous phrase (e.g. “**brakes**” transcribed as “**breaks**”). More common were mistranscriptions that shared a similar syllabic structure and some (but not all) speech sounds with the reference word(s) (e.g. “**catchment**” transcribed as “**Catharine**”). These findings align with the way in which ASR works; the system uses technical characteristics of the acoustic signal to determine the speech sounds produced. Issues occur when the system incorrectly maps those speech sounds to the words spoken, which results in substitutions that sound similar to the spoken content.

In our analysis of substitution errors, we found that just over 25% of substitutions resulted in errors that (i) do not retain the original meaning but (ii) make sense and (iii) are a plausible interpretation of the spoken content. This is particularly pertinent with regard to transcript checking procedures. Given that outputs may be viewed as "broadly accurate" or "good enough" there is a risk of increasingly cursory reviewing processes (Preston et al. 2026: 4). Additionally, if the review involves reading an ASR transcript while listening to the audio recording, this will impact the listener's perception of the speech content (e.g. Fraser & Kinoshita 2021). Findings from a pilot study conducted by the first author (Harrington & Tompkinson 2026) reveal that transcript reviewers did not consistently identify all errors in ASR transcripts despite favourable audio conditions. Incorrectly transcribed place names and pronouns posed particular problems, being corrected by less than 50% of participants. More testing on whether listeners are significantly primed by ASR transcripts is therefore an important area for future research.

In this study, reference transcripts were produced by correcting the output of CrisperWhisper. We found that several errors produced by the ASR system were missed during initial rounds of review and, as a result, spent a substantial amount of time producing several drafts of each transcript. We found that, in our case, extensive phonetic training and an understanding of speech perception did not necessarily mitigate the impacts of priming. This highlights the challenge of reviewing ASR outputs, which is not the same process as producing a transcript from scratch.

The approach to substitution analysis used in this study (i.e. binary classifications of meaning retention and nonsensicality) was a relatively subjective process. A more systematic manner of evaluating these factors should be developed bearing in mind that errors will always be context-dependent. Any evaluation applies only to the data on which it has been carried out and the system it has tested (e.g. van der Vloed 2026). Thus, an ideal approach would entail the evaluation of ASR outputs for a specific use case, in collaboration with the organisations looking to use the system, testing on realistic spoken interactions and involving practitioners in the evaluation process. This would also enable the incorporation of appropriate human-based evaluations enabling both a more nuanced and comprehensive evaluation.

## *Conclusion*

Linguistic evaluations of ASR performance can offer several tangible benefits, such as the diagnosis of regional biases (Serditova et al. 2025) and the development of more equitable ASR systems (Wassink et al. 2022). In this study, we employed domain-specific knowledge to interpret the content of ASR transcripts and investigate what types of errors these systems make, what types of words those errors impact, and how those errors affect the content of the transcript. We find that there are several risks associated with the use of ASR for the transcription of speech in high-stakes contexts. First, all ASR transcripts in our study contained substantial errors which often affected words like proper nouns and pronouns, the mistranscription of which could have serious implications in, for example, police investigations. Even the transcript with the lowest number of errors contained two incorrectly transcribed place names. Second, many of the substitutions in the transcripts both made sense within the given context and were phonetically similar. We suggest that these errors would likely be very challenging to identify within a human verification process. This raises concerns about the idea that human verification is an appropriate safeguard to ensuring that errors within ASR outputs are captured before impacting real-world decisions.

There is widespread excitement about the implementation of AI for time-consuming 'administrative' processes like transcription, particularly within the UK public sector (Government Digital Service 2025). Evaluations tend to centre on time-saving measures, with little consideration given to the quality of the outputs and, crucially, the efficacy and appropriateness of safeguarding procedures. When decision-makers are presented with opportunities to procure AI speech technologies, it is likely that providers cite overall metrics like accuracy percentages or word error rate. We have demonstrated in this paper what actually occurs in the inaccurate sections of ASR transcripts. While some errors do not have a meaningful impact on the transcript, there are a substantial proportion of errors that result in nonsense or meaning shifts. Practitioners and policymakers should understand how ASR systems function and perform, and be aware of the risks associated with their use on high-stakes materials.

### *Endnotes*

1. Statistics were extracted from the Open ASR Leaderboard on 25th November 2026.
2. Note that since conducting this analysis, the developers have released CrisperWhisper 2.0 (Wagner 2026).

### *Acknowledgements*

This work was funded by an International Association for Forensic Phonetics and Acoustics grant. The Viking cluster, a high performance compute facility provided by the University of York, was used to run CrisperWhisper. We are grateful for computational support from Philip Harrison, and feedback on drafts by James Tompkinson,Vincent Hughes and Amelia Gully.

### *References*

Atwany, H., Waheed, A., Singh, R., Choudhury, M., & Raj, B. 2025. Lost in transcription, found in distribution shift: Demystifying hallucination in speech foundation models. *Findings of the Association for Computational Linguistics: ACL 2025*, 23181–23203.

Barański, M., Jasiński, J., Bartolewska, J., Kacprzak, S., Witkowski, M., & Kowalczyk, K. 2025. Investigation of Whisper ASR hallucinations induced by non-speech audio. *ICASSP 2025-2025 IEEE International Conference on Acoustics, Speech and Signal Processing (ICASSP)*, 1–5.

Boersma, P., & Weenink, D. 2025. *Doing phonetics by computer* [Computer program] (Version 6.4.27). http://www.praat.org/

Bucholtz, M. 2000. The politics of transcription. *Journal of Pragmatics, 32*(10), 1439-1465.

College of Policing. 2026. Deciding when to use AI. https://www.college.police.uk/guidance/building-ai-enabled-tools-and-systems/deciding-when-use-ai

Dua, M., Akanksa, & Dua, S. 2023. Noise robust automatic speech recognition: review and analysis. *International Journal of Speech Technology, 26,* 475–519.

Ernestus, M. & Warner, N. 2011. An introduction to reduced pronunciation variants. *Journal of Phonetics, 39*, 253–260.

Feng, S., Kudina, O., Halpern, B. M., & Scharenborg, O. 2021. Quantifying bias in automatic speech recognition. *arXiv preprint arXiv:2103.15122.*

Fraser, H. 2022. A framework for deciding how to create and evaluate transcripts for forensic and other purposes. *Frontiers in Communication, 7*, 898410.

Fraser, H., & Kinoshita, Y. 2021. Injustice arising from the unnoticed power of priming: How lawyers and even judges can be misled by unreliable transcripts of indistinct forensic audio. *Criminal Law Journal, 45*(3), 142–152.

Government Digital Service. 2025. Artificial Intelligence Playbook for the UK Government. Department for Science, Innovation and Technology. https://assets.publishing.service.gov.uk/media/67aca2f7e400ae62338324bd/AI_Playbook_for_the_UK_Government__12_02_.pdf

Green, B. 2022. The flaws of policies requiring human oversight of government algorithms. *Computer Law & Security Review, 45*, 105681.

Hain, T., Woodland, P. C., Evermann, G., Gales, M. J. F., Liu, X., Moore, G. L., Povey, D., & Wang, L. 2005. Automatic transcription of conversational telephone speech. *IEEE Transactions on Audio, Speech, and Language Processing, 13*(6), 1173–1185.

Harrington, L. & Tompkinson, J. 2026, July 8–10. *Assessing the accuracy of a human post-editing approach to AI-generated transcripts of interview recordings* [Conference presentation]. The International Investigative Interviewing Research Group Conference 2026, Leuven, Belgium.

Harrington, L. 2023. Incorporating automatic speech recognition methods into the transcription of police-suspect interviews: factors affecting automatic performance. *Frontiers in Communication, 8*, 1165233.

Harrington, L. 2024. *Towards improving transcripts of audio recordings in the criminal justice system* (Doctoral dissertation, University of York).

Harrington, L., Hughes, V., Harrison, P., Foulkes, P., Wormald, J., Kelly, F., & van der Vloed, D. 2025. Variability in performance across four generations of automatic speaker recognition systems. *Proceedings of Interspeech 2025*, 3993–3997.

Haworth, K. 2018. Tapes, transcripts and trials: The routine contamination of police interview evidence. *The International Journal of Evidence & Proof, 22*(4), 428–450.

He, X. & Whitehall, J. 2026. Survey of end-to-end multi-speaker automatic speech recognition for monaural audio. *Computer Speech & Language, 99*, 101925.

Hughes, V., Xu, C., Foulkes, P., Harrison, P., Welch, P., Kelly, F. & Van Der Vloed, D. 2024. Exploring individual speaker behaviour within a forensic automatic speaker recognition system. *Proceedings of The Speaker and Language Recognition Workshop (Odyssey 2024)*.

Jurafsky, D., Bell, A., Gregory, M., & Raymond, W. 2001. Probabilistic relations between words: Evidence from reduction in lexical production. In J. L. Bybee & P. J. Hopper (Eds.), *Frequency and the emergence of linguistic structure* (pp. 229–254).

Jurafsky, D., & Martin, J. H. 2026. *Speech and Language Processing: An Introduction to Natural Language Processing, Computational Linguistics, and Speech Recognition with Language Models* (3rd ed. draft).

Kim, S., Arora, A., Le, D., Yeh, C.-F., Fuegen, C., Kalinli, O., & Seltzer, M.L. 2021. Semantic Distance: A New Metric for ASR Performance Analysis Towards Spoken Language Understanding. *Proceedings of Interspeech 2021*, 1977–1981.

Klering, E. H., Cortes, E. G., Chernenko, T., Trarbach, M. V., Ramos, G. d. O., Rigo, S. J., Dupont, M., Vianna, A. L. T., Santos, G. K. d., Pereira, V. M., Araujo, D. A. d., & Kunst, R. 2026. A Dataset for Evaluating ASR on Specialized Vocabulary. *Proceedings of the 15th Language Resources and Evaluation Conference*, 470–480.

Koenecke, A., Nam, A., Lake, E., Nudell, J., Quartey, M., Mengesha, Z., Toups, C., Rickford, J. R., Jurafsky, D., & Goel, S. 2020. Racial disparities in automated speech recognition. *Proceedings of the National Academy of Sciences of the United States of America, 117*(14), 7684–7689.

Lange, N. D., Thomas, R. P., Dana, J., & Dawes, R. M. 2011. Contextual biases in the interpretation of auditory evidence. *Law and Human Behavior, 35*(3), 178–187.

Markl, N. 2022. Language variation and algorithmic bias: understanding algorithmic bias in British English automatic speech recognition. *2022 ACM Conference on Fairness, Accountability, and Transparency*, 521–534.

McGuire, M., & Larson-Hall, J. 2025. Assessing Whisper automatic speech recognition and WER scoring for elicited imitation: Steps toward automation. *Research Methods in Applied Linguistics, 4*(1), 100197.

Ministry of Justice. 2025. *AI action plan for justice.* https://www.gov.uk/government/publications/ai-action-plan-for-justice/ai-action-plan-for-justice#use-speech-and-translation-ai-to-free-frontline-staff-from-admin-burdens

Ministry of Justice and HM Prison and Probation Service. 2026. *Justice Transcribe*. https://www.gov.uk/government/publications/justice-transcribe

Nagy, N. & Sharma, D. 2013. Transcription. In R. J Podesva & D. Sharma (Eds.), *Research Methods in Linguistics* (pp. 235-256). Cambridge University Press.

National Audit Office. 2026. Productivity Improvement: The Justice Transcribe AI-enabled Transcription Tool. *HC 615 (Session 2026–27).*

Nayeem, M., Tabrej, M. S., Deb, K. J., Goswami, S., & Hakim, M. A. 2025. Automatic Speech Recognition in the Modern Era: Architectures, Training, and Evaluation. *arXiv preprint arXiv:2510.12827.*

OpenAI. 2023, 25 September. ChatGPT can now see, hear, and speak. *OpenAI.* https://openai.com/index/chatgpt-can-now-see-hear-and-speak/

Preston, K., Habli, I., Lawton, T., Hayat, Y., Cunningham, A. & Sujan, M. 2026. The Myth of Effortless Clinical Documentation Automation: A Critical Sociotechnical Commentary on Ambient Digital Scribes. *Journal of Medical Internet Research, 28*:e109717.

Radford, A., Kim, J. W., Xu, T., Brockman, G., McLeavey, C., & Sutskever, I. 2023. Robust speech recognition via large-scale weak supervision. *Proceedings of the 40th International Conference on Machine Learning*, PMLR 202:28492–28518.

Reid, S. A., Abeyta, S., Piza, E. L., Connealy, N. T., & Sytsma, V. A. 2026. AI vs. human transcription: evaluating accuracy and meaning in police body-worn camera footage. *Journal of Experimental Criminology*, 1–22.

Sargeant, H. 2026. *Artificial intelligence and justice – an evidence scoping review.* Nuffield Foundation.

Serditova, D., Tang, K., Steffens, J. 2025. Automatic Speech Recognition Biases in Newcastle English: an Error Analysis. *Proceedings of Interspeech 2025*, 3204–3208.

Seuren, L.M., Williams, R. & Cresswell, K. 2026. Beyond productivity: a NASSS-informed review of implementation risks and research priorities for ambient AI scribes in healthcare. *BMJ Digital Health & AI*, e000091.

Sohoglu, E., Peelle, J. E., Carlyon, R. P., & Davis, M. H. 2014. Top-down influences of written text on perceived clarity of degraded speech. *Journal of Experimental Psychology: Human Perception and Performance, 40*(1), 186-199.

Srivastav, V., Zheng, S., Bezzam, E., Bihan, E.L., Koluguri, N.R., Żelasko, P., Majumdar, S., Moumen, A., Gandhi, S. 2026. Open ASR Leaderboard: Towards Reproducible and Transparent Multilingual and Long-Form Speech Recognition Evaluation. *Proc. Interspeech 2026*, 6692–6697.

Tuttösí, P., Dhillon, M., Sang, L., Eastwood, S., Bhatia, P., Dinh, Q.M., Kapoor, A., Jin, Y. & Lim, A. 2026. BERSting at the screams: A benchmark for distanced, emotional and shouted speech recognition. *Computer Speech & Language, 95*, 101815.

Van der Vloed, D. 2026. Data selection in forensic automatic speaker comparison. *Forensic Science International,* 113093.

Wagner, L., Thallinger, B., & Zusag, M. 2024. CrisperWhisper: Accurate Timestamps on Verbatim Speech Transcriptions. *Proceedings of Interspeech 2024*, 1265–1269.

Wagner, L. 2026. Transcription Policy as a Latent Variable: Activating Controllable Verbatim ASR with Word-Level Timing. *Proceedings of Interspeech 2026*, 650–659.

Walley, P. & Glasspoole-Bird, H. 2025. An Evaluation of the Pilot Application of Artificial Intelligence to Witness Statement and Report Generation at Hertfordshire Constabulary. *Centre for Policing Research and Learning, The Open University.*

Wang, D., Wang, X. & Lv, S. 2019. An Overview of End-to-End Automatic Speech Recognition. *Symmetry, 11*(8), 1018.

Warren, R. M. 1970. Perceptual Restoration of Missing Speech Sounds. *Science, New Series 167*(3917), 392-393.

Wassink, A. B., Gansen, C., & Bartholomew, I. 2022. Uneven success: automatic speech recognition and ethnicity-related dialects. *Speech Communication, 140*, 50–70.

Wormald, J. 2016. *Regional variation in Panjabi-English* (Doctoral dissertation, University of York).

Wormald, J. & Harrington, L. 2026. Investigating the effect of spontaneous speech features on by-speaker automatic speech recognition performance, *arXiv preprint arXiv:xxxx.xxxxx.*

Zhang, T., Kishore, V., Wu, F., Weinberger, K. Q. & Artzi, Y. 2019. BERTScore: Evaluating text generation with BERT. *arXiv preprint arXiv:1904.09675*.